\documentclass{article}
\usepackage{spconf,amsmath,graphicx}
\usepackage{booktabs}
\usepackage{subcaption}
\usepackage{amssymb}
\usepackage{comment}
\usepackage{url}

\usepackage{algpseudocode}
\usepackage[ruled,vlined]{algorithm2e}
\newcommand{\set}[1]{\mathcal{#1}}
\newcommand{\vect}[1]{\mathbf{#1}}

\newcommand{\myparagraph}[1]{\noindent \textbf{#1}}

\title{Exploring Active Learning for Semiconductor Defect Segmentation}

\name{Lile Cai, Ramanpreet Singh Pahwa, Xun Xu, Jie Wang, Richard Chang, Lining Zhang, Chuan-Sheng Foo \thanks{This research is supported by the Agency for Science, Technology and Research (A*STAR) under its AME Programmatic Funds (Grant No. A20H6b0151) and Career Development Fund (Grant no. C210812046).}}
\address{Institute for Infocomm Research (I$^2$R), A*STAR, Singapore.\\
\{caill,ramanpreet\_pahwa,xu\_xun,wang\_jie,richard\_chang,zhang\_lining,foo\_chuan\_sheng\}@i2r.a-star.edu.sg.}

\begin{document}
%
\maketitle
\begin{abstract}
The development of X-Ray microscopy (XRM) technology has enabled non-destructive inspection of semiconductor structures for defect identification. Deep learning is widely used as the state-of-the-art approach to perform visual analysis tasks. However, deep learning based models require large amount of annotated data to train. This can be time-consuming and expensive to obtain especially for dense prediction tasks like semantic segmentation. In this work, we explore active learning (AL) as a potential solution to alleviate the annotation burden. We identify two unique challenges when applying AL on semiconductor XRM scans: large domain shift and severe class-imbalance. To address these challenges, we propose to perform contrastive pretraining on the unlabelled data to obtain the initialization weights for each AL cycle, and a rareness-aware acquisition function that favors the selection of samples containing rare classes. We evaluate our method on a semiconductor dataset that is compiled from XRM scans of high bandwidth memory structures composed of logic and memory dies, and demonstrate that our method achieves state-of-the-art performance. 
\end{abstract}
\begin{keywords}
Active Learning, Semantic Segmentation, Semiconductor Structures
\end{keywords}
\section{Introduction}
\label{sec:intro}
\begin{figure}[ht]
	\centering
	\includegraphics[width=\linewidth]{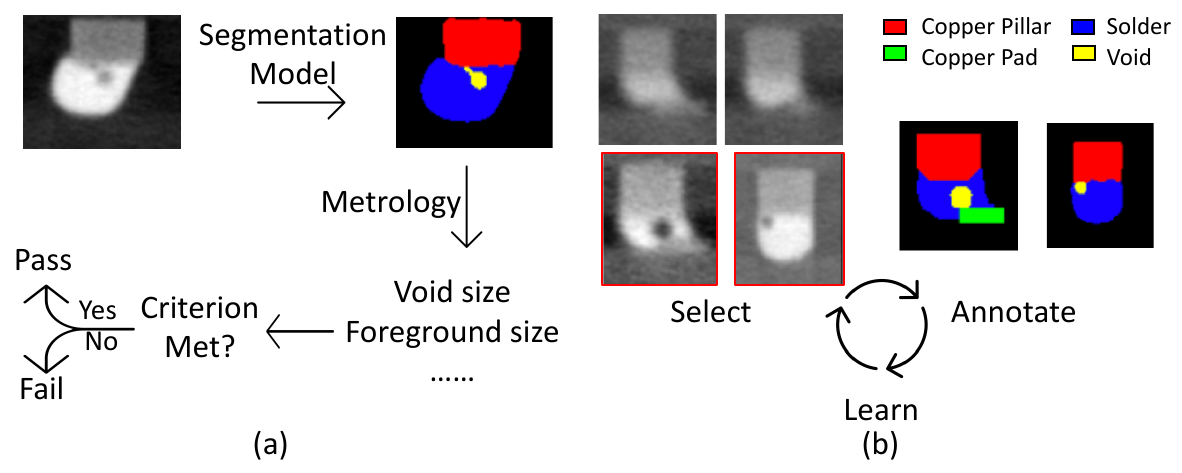}
	\vspace{-0.5cm}
	\caption{Problem statement. (a) Semantic segmentation facilitates automatic defect identification for semiconductor structures. (b) Active learning offers a potential solution to alleviate the annotation burden for learning deep models.}
	\label{fig:intro}
	\vspace{-0.5cm}
\end{figure}

The development of X-Ray microscopy (XRM) technology has enabled non-destructive techniques (NDT) applications in inspection of semiconductor structures. Facilitated by machine learning and sophisticated image processing, it is now possible to automatically identify important structures in semiconductor XRM scans. A use case is illustrated in Fig.~\ref{fig:intro}a, where segmentation technique is employed to segment different regions in the XRM scan and metrology information is extracted from the segmentation results. The structure can be classified as either ``pass" or ``fail" depending on whether some predefined criterion is met (e.g., if the ratio of void over foreground is above a certain pre-decided threshold, the structure is detected to be defective). In this work, we focus on the task of semantic segmentation for semiconductor structures.

Deep learning (DL) is the state-of-the-art technique for visual recognition tasks. Attempts have been made to apply deep learning-based models on the XRM scans of semiconductor and results have been promising \cite{eptc_20, pahwa2021automated, ectc_22}. However, previous work \cite{eptc_20, pahwa2021automated} focuses on designing specific deep learning models for semiconductor structures, and large amount of labelled data is needed to train the model. The laborious and costly process of data annotation hinders the application of DL in semiconductor manufacturing. In this work, we explore active learning (AL) as a potential technique to mitigate the annotation burden. AL attempts to maximize a model's performance while annotating the fewest samples possible. It is typically an iterative process, where in each cycle, an acquisition function is used to select a set of informative samples, and the selected samples are sent to an oracle for annotation. The model is then re-trained on all the samples annotated so far and the process iterates until the annotation budget is exhausted or satisfactory performance is achieved. The process is illustrated in Fig.~\ref{fig:intro}b.

There are some unique challenges when applying AL to semiconductor data. First, there is a large domain shift between the XRM scans and natural images (e.g. ImageNet). This matters because segmentation models are usually initialized with ImageNet pretrained weights, and the large domain shift may affect the effectiveness of the ImageNet pretrained weights, especially during the early cycles of AL when the labelling budget is low. Second, the semiconductor data exhibits severe class-imbalance, e.g., the void class is rare and occupies a small area within an image. Deep learning models are trained by back-propagating the loss on all samples and performance on minority classes can degrade when the gradients are dominated by data from the majority classes. In this work, we propose to perform contrastive pretraining and rareness-aware selection to address these challenges. Our contributions can be summarized as below:
\begin{itemize}
\vspace{-0.1cm}
    \item We propose to employ contrastive pretraining on the semiconductor dataset and to use the contrastive pretrained weights for model initialization at each AL cycle. We demonstrate that this significantly outperforms initialization with ImageNet pretrained weights.
\vspace{-0.1cm}
    \item We propose a rareness-aware acquisition function that favors the selection of samples containing minority classes to address the class-imbalance issue in semiconductor data. We benchmark the proposed method against state-of-the-art AL methods and demonstrate that our method outperforms others on the semiconductor data.
\end{itemize}

\section{Related Work}
\label{sec:related_work}
\myparagraph{Active Learning}
Based on the criterion used to query samples, AL methods can be broadly categorized into uncertainty-based, diversity-based and hybrid methods. Uncertainty-based methods select samples that the current model is most uncertain about to label. Ensemble-based method \cite{beluch2018power} has shown to provide more calibrated uncertainty estimation. Yoo and Kweon  \cite{yoo2019learning} proposed a task-agnostic method to estimate sample uncertainty by employing a loss prediction module. Diversity-based methods aim to select a diverse yet representative set of samples to label. CoreSet \cite{sener2017active} selects a set of samples that minimize the difference between the average empirical training loss on this subset and the average empirical loss on the entire dataset. CoreGCN \cite{caramalau2021sequential} extends CoreSet to operate on features learned by graph convolutional network. VAAL \cite{sinha2019variational} trains an auto-encoder in an adversarial manner and uses the discriminator score to select samples that are most different from already labelled ones. Hybrid methods combine both uncertainty and diversity in selecting samples to label. BADGE \cite{ash2019deep} applies k-means++ on the gradient embedding of samples. The gradient embedding is computed as the output of the penultimate layer of the network scaled by prediction confidence and thus captures both uncertainty and diversity signals. In this work, we propose a rareness-aware acquisition function that not only considers uncertainty and diversity, but also the rareness of a sample.

\myparagraph{Contrastive Learning}
Contrastive learning is an unsupervised learning technique that learns representations by increasing the similarity of representations of positive sample pairs and pushing apart those of negative sample pairs. Methods for contrastive learning differ in how they define the sample pairs. Positive pairs are typically formed by two augmented views of the same image and negative pairs are formed by different images. SimCLR \cite{chen2020simple} treats other samples in the current batch as negative, while MoCo \cite{he2020momentum} maintains negative samples in a queue. SimSiam \cite{chen2021exploring} eliminates the need for negative samples by applying a stop-gradient operation to Siamese networks. The above methods are developed for classification models. In this work, we adapt SimCLR to perform contrastive learning for segmentation models.


\myparagraph{Unsupervised Pretraining for Active Learning} Unsupervised learning has been explored as a pretraining technique to leverage unlabelled data in active learning. The pioneering work \cite{simeoni2021rethinking} proposed to perform clustering-based pretraining on all data once and use the learned weights to initialize model at each AL cycle. A similar approach was adopted in \cite{gudovskiy2020deep}, where the unsupervised learning signal is given by a rotation prediction pretext task. Both works only studied image classification; different from previous work, we focus on the more challenging semantic segmentation task.

\section{Method}
\label{sec:method}
In this section, we first introduce our method to perform contrastive pretraining with segmentation models, followed by description on how we perform rareness-aware sampling to select samples for annotation. 

\subsection{Contrastive Pretraining for Segmentation Models}
The loss function in contrastive learning measures the similarities of sample pairs in a feature space. A commonly used loss function called InfoNCE \cite{van2018representation} is defined as:
\begin{equation}
\mathcal{L}_{CL} = \frac{1}{N}\sum\limits_{i=1}^N -\log\frac{\exp(\vect{v}_i\cdot \vect{v}_{i}^{+}/\tau)}
{\exp(\vect{v}_i\cdot \vect{v}_{i}^{+}/\tau) + \sum\limits_{\vect{v}_{i}^{-}\in \set{V}^{-}}\exp(\vect{v}_i\cdot \vect{v}_{i}^{-}/\tau)},
\label{eq:L_CL}
\end{equation}
where $\tau$ is a temperature hyper-parameter, $\vect{v}_{i}$ is a feature vector for sample $i$, $\vect{v}_{i}^{+}$ is the feature vector of a positive sample of instance $i$ that is typically generated by applying data augmentation to the input image, and $\set{V}^{-}$ is a set of negative samples that are randomly drawn from training samples excluding $i$. No labels are involved in the computation of $\mathcal{L}_{CL}$. The constrastive loss learns meaningful features by encouraging the feature representation of positive pairs to be similar, while pushing features of negative pairs apart.

\begin{figure}[ht]
	\centering
	\includegraphics[width=\linewidth]{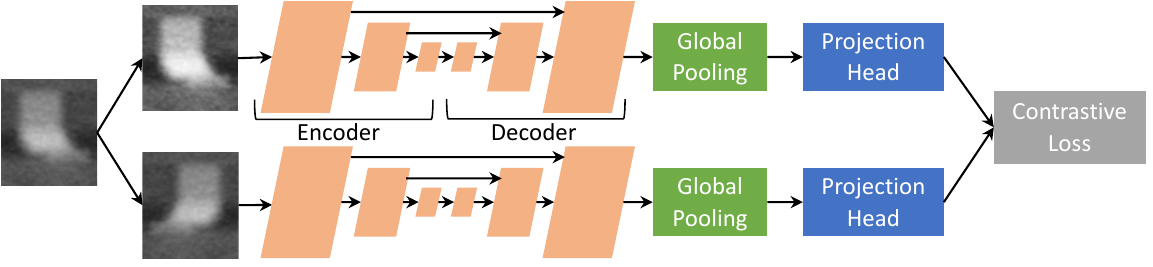}
	\caption{The proposed method of performing contrastive learning with segmentation model.}
	\label{fig:cl_for_segmentation}
\end{figure}

Our method of performing contrastive learning with segmentation models is illustrated in Fig.~\ref{fig:cl_for_segmentation}. The structure of a segmentation model typically consists of an encoder, a decoder and segmentation head.
We apply a global pooling layer to the decoder output to produce a feature vector for sample $i$. Following the design of SimCLR \cite{chen2020simple} and MoCo v2 \cite{chen2020improved}, we attach a 2-layer MLP projection head to the feature vector to obtain the final $\vect{v}_{i}$; this $\vect{v}_{i}$ is then used to compute $\mathcal{L}_{CL}$ defined in Eq.~(\ref{eq:L_CL}). The model is trained from scratch by minimizing $\mathcal{L}_{CL}$ on the entire unlabelled training set. After contrastive pretraining, we use the learned parameters from layers before the global pooling layer to initialize the segmentation model during each AL cycle. 
\subsection{Rareness-Aware Acquisition Functions}

XRM scans of semiconductor components typically exhibit severe class-imbalance, e.g., the void class (corresponding to defects) is rare and occupies a small area in an image. Motivated by the observation that labelling more samples from rare classes improves the performance of deep learning model on class-imbalanced datasets \cite{cai2021revisiting}, we propose to employ ``rareness" as a criterion to select samples for active learning. Our rareness measure is based on estimating the class distribution using pseudo labels. Let $x$ denote a pixel and $M_{t-1}$ the segmentation model trained in the previous AL cycle. The pseudo label $\hat{y}$ for $x$ is given by: $\hat{y}(x) = \arg\max_{c \in \set{C}} p(y=c|x, M_{t-1})$, where $\set{C}$ is the set of class labels. This gives the posterior of class distribution $p(c)$ as: $ p(c) =|\{x \mid \hat{y}(x) = c \land x \in \set{X}\}|/|\set{X}|$, where $\set{X}$ is the set of pixels in the training set. The rareness score of pixel $x$ is then defined as:
\begin{equation} 
    r(x) = e^{-p(\hat{y}(x))}.
\label{eq:rareness_score}
\end{equation}
The rareness score of an image $I$ is obtained by aggregating the pixel-wise scores for pixels in the image:
\begin{equation} 
    r(I) = f_{aggr}(r(x)), x \in I.
\label{eq:aggr}
\end{equation}
We complement the rareness score with uncertainty and diversity scores: 
\begin{equation}
s(I) = r(I) + u(I) + d(I, \set{L}), 
\label{eq:af}
\end{equation}
where $u(I)$ is the uncertainty score for image $I$, $d(I, \set{L})$ is a diversity score that measures the distance between $I$ and the set of previous selected samples $\set{L}$. The uncertainty score $u(I)$ is defined as:
\begin{equation} 
    u(I) = f_{aggr}(u(x)),  x \in I,
\label{eq:u_I}
\end{equation}
where $u(x) = -\sum_{c \in \set{C}} p(y=c|x)\log p(y=c|x)$ is the entropy of the predictive posterior. The distance between image $I$ and $\set{L}$ is defined as:
$d(I, \set{L}) = \min_{S \in \set{L}} ||\vect{f}_I - \vect{f}_S||_2$,  
where $\vect{f}$ is a feature vector for an image that is computed by average pooling of the decoder output.

During each cycle of AL, we select samples that maximize Eq.~(\ref{eq:af}) greedily until the annotation budget is met. The greedy algorithm is summarized in Algorithm~\ref{greedy_algo}. We use $max(\cdot)$ for $f_{aggr}(\cdot)$ in Eqs.~(\ref{eq:aggr}) and (\ref{eq:u_I}), and investigate the effect of different aggregation methods in Section~\ref{sec:exp_res}.

\begin{algorithm}
\SetAlgoLined
\SetKwInOut{Input}{Input}
\SetKwInOut{Output}{Output}
\Input{labelled set of $\set{L}_{t-1}$, unlabelled set   $\set{U}_{t-1}$, budget $K$ for cycle $t$}
\Output{selected set $\set{B}_{t}$}

$\set{B}_t=\emptyset$\;
$\set{U}_t=\set{U}_{t-1}$\;
\While{$|\set{B}_t|<K$}
    {
        $\hat{I}=\arg\max\limits_{I\in\set{U}_t} [r(I) + u(I) + d(I, \set{L}_{t-1}\cup \set{B}_t )$]\;
        $\set{B}_t=\set{B}_t\cup \hat{I}$\;
        $\set{U}_t=\set{U}_t \setminus \hat{I}$\;
       
    }
$\set{L}_t=\set{L}_{t-1}\cup \set{B}_t$\;
    
\caption{Greedy Active Selection}\label{greedy_algo}
\end{algorithm}

\section{Experiments}
\label{sec:experiments}
\subsection{Experimental Setup}
\myparagraph{Datasets} Our dataset is compiled from 3D XRM scans of high bandwidth memory structures composed of logic and memory dies. The logic die consists of three classes, namely, copper pillar, solder and void; the memory die contains one additional class named copper pad. The dataset contains 25 3D scans for logic die, and 53 3D scans for memory die. We project the 3D scans to coronal view and slice each scan into 48 to 82 2D images. The width of the images is in the range [51,96], and the height is in the range [57,96]. We split the dataset into training (80\%)/testing (20\%) sets at the 3D scan level to avoid data leakage, resulting in 4,086 and 964 images for training and testing respectively. We perform contrastive pretraining and active learning on the training split and report the performance of the trained model on the testing split.

\myparagraph{Segmentation Model} We use a U-Net \cite{ronneberger2015u} with ResNet-18 \cite{he2016deep} backbone. During each AL cycle, the model is trained with RMSprop optimizer with weighted cross entropy loss. The weight for each class is set inversely to the class frequency in current labelled data. Hyper-parameters are set as follows: number of epochs = 50, learning rate = 1e-4, which is reduced to 1e-5 after 25 epochs, batch size = 16, weight decay = 1e-8, momentum = 0.9. For data augmentation, the image is first resized by a factor randomly selected in \{0.5, 0.75, 1.0, 1.25, 1.5\}, and then randomly cropped and padded to a fixed size of $96\times96$. Horizontal flipping and vertical flipping are randomly applied with probability 0.5. We use the open source library Segmentation Models Pytorch \cite{Yakubovskiy2019}. 

\myparagraph{Contrastive Pretraining} We use the implementation of an open source library \cite{jiang2021self} for SimCLR. The model is trained for 2000 epochs with batch size 256. The learning rate is 0.5 with cosine schedule. For augmentation, images are cropped and resized to a fixed size of $64\times64$, and we add random vertical flipping and remove random conversion to gray scale. Other augmentation techniques and hyper-parameters are kept the same as used in \cite{jiang2021self}. It takes on average 3.5 hours to perform pretraining on the semiconductor training split with one Tesla V100 GPU.

\subsection{Experimental Results}
\label{sec:exp_res}
\begin{figure}[t]
\centering
\hspace{-1cm}
\begin{subfigure}{0.27\textwidth}
	\includegraphics[width=\linewidth]{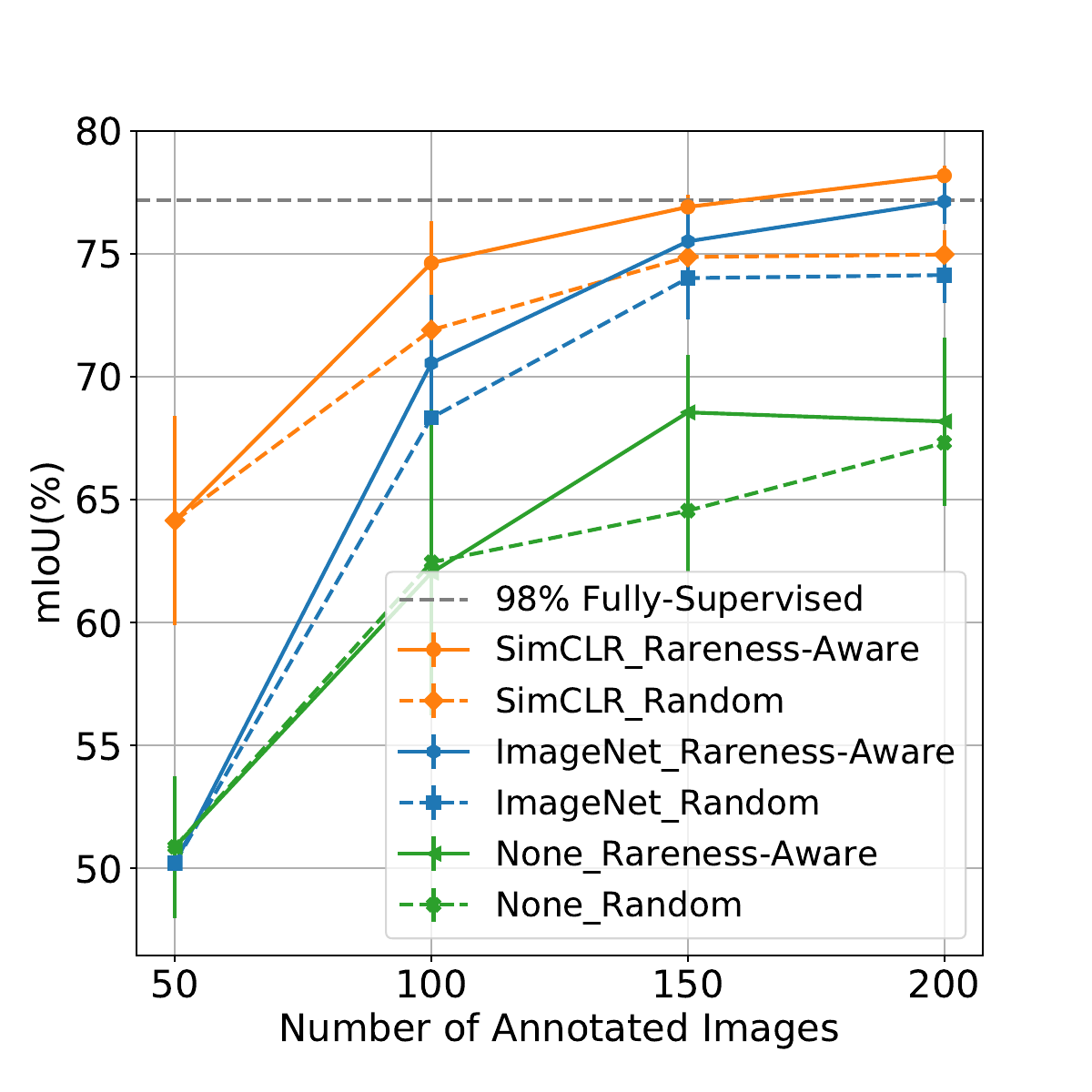}
	\vspace{-0.5cm}
	\caption{}
    \label{fig:effect_pretrained_weights}
\end{subfigure}
\hspace{-0.4cm}
\begin{subfigure}{0.27\textwidth}
	\includegraphics[width=\linewidth]{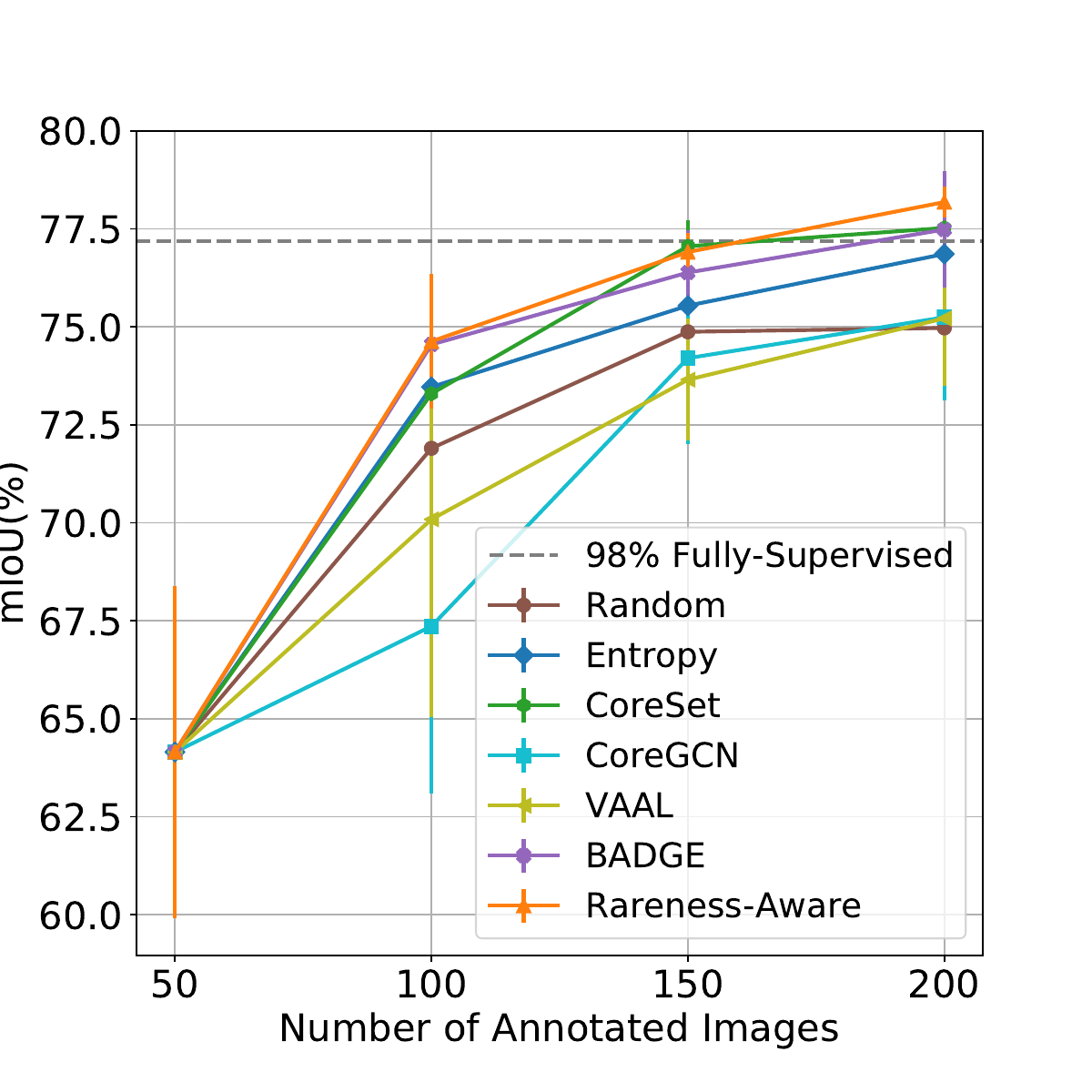}
	\vspace{-0.5cm}
	\caption{}
	\label{fig:effect_al_selection}
\end{subfigure}
\hspace{-1cm}
\vspace{-0.3cm}
\caption{Results on semiconductor XRM dataset. (a) Effect of pretrained weights. (b) Effect of AL selection strategy. Each point and error bar represent the mean and standard deviation of 5 runs, respectively.}
\vspace{-0.5cm}
\end{figure}

\myparagraph{Effect of Pretrained Weights} The effect of initialization with pretrained weights on active learning is shown in Fig.~\ref{fig:effect_pretrained_weights}. We fix the selection strategies to be Random and Rareness-Aware. The SimCLR pretrained weights outperform ImageNet pretrained weights by a significant margin, especially at the early stage of AL. We also observe that models perform poorly when randomly initialized (None\_Random and None\_Rareness-Aware in Fig.~\ref{fig:effect_pretrained_weights}). With 200 images ($\sim$4.9\% of the entire training set), our method (\mbox{SimCLR\_Rareness-Aware}) achieves 78.18\% mIoU, which is 98\% of the performance obtained when the entire training set is annotated.

\myparagraph{Effect of AL Selection Strategy}
We compare our rareness-aware acquisition function with other baselines (Random and Entropy) and state-of-the-art AL methods (CoreSet \cite{sener2017active}, CoreGCN \cite{caramalau2021sequential}, VAAL \cite{sinha2019variational} and BADGE \cite{ash2019deep}) in Fig.~\ref{fig:effect_al_selection}. For fair comparison, all competing methods start from the same first batch that is randomly selected and use the same SimCLR pretrained weights for initialization. Our method consistently outperforms other methods at different labeling budgets.

\myparagraph{Ablation Studies}
Ablation study on the three terms of our rareness-aware acquisition function is provided in Table~\ref{tab:ablation_rare}. The rareness term improves mIoU by 1.06\% over Entropy, and 0.19\% over Entropy+Feature. This demonstrates the effectiveness of the proposed rareness term. The effect of the aggregation function on the rareness-aware acquisition function is shown in Table~\ref{tab:effect_aggre}. Using Max gives better performance than Mean; this could be because the rare class \emph{void} only occupies a very small area in an image and will contribute much less to the aggregated score if using Mean than using Max.

\begin{table}
\small

\begin{subtable}{0.45\textwidth}
\centering
\begin{tabular}{ccccr}
\toprule
Entropy    & Feature   & Rareness    & mIoU(\%) \\
\midrule

\checkmark &            &            &  76.86 (1.44)   \\
\checkmark &            & \checkmark &  77.92 (0.79)   \\
\checkmark & \checkmark &            &  78.00 (0.65)   \\
\checkmark & \checkmark & \checkmark &   \textbf{78.19} (0.40)   \\
\bottomrule
\end{tabular}
\caption{}
\label{tab:ablation_rare}
\end{subtable}

\begin{subtable}{0.45\textwidth}
\centering
\begin{tabular}{crrr}
\toprule
Budget    & 100   & 150    & 200 \\
\midrule
Mean      &  74.23 (1.48)  & 76.31 (1.53)            &  77.29 (0.88)   \\
Max       & \textbf{74.63} (1.71)  &  \textbf{76.91} (0.49)   & \textbf{78.19} (0.40)\\
\bottomrule
\end{tabular}
\caption{}
\label{tab:effect_aggre}
\end{subtable}
\vspace{-0.3cm}
\caption{Ablation studies for rareness-aware acquisition function. (a) Ablation on individual terms of rareness-aware acquisition function at budget=200. (b) Effect of aggregation function $f_{aggr}$ for rareness-aware acquisition function. We report the mean and standard deviation (in brackets) of 5 runs.}
\label{tab:ablation_study}
\vspace{-0.5cm}
\end{table}

\begin{figure}[ht]

	\centering
	\includegraphics[width=\linewidth]{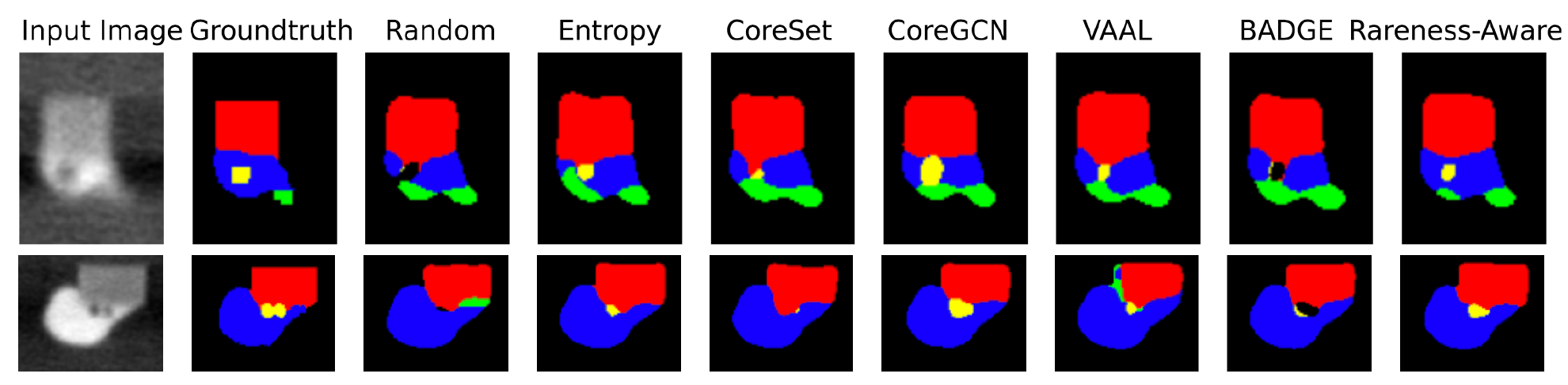}
	\caption{Visualization of segmentation results for a memory die (top row) and a logic die (bottom row) at budget=100.}
	\label{fig:vis_benchmark}
	\vspace{-0.5cm}
\end{figure}

\myparagraph{Qualitative Results}
We present segmentation results of models trained by images selected by different AL strategies in Fig.~\ref{fig:vis_benchmark}. Our method (Rareness-Aware) is able to segment the void (in yellow) well while other methods either miss the detection (e.g., CoreSet, BADGE) or fail to delineate the shape of the void accurately (e.g., CoreGCN, VAAL).

\section{Conclusions}
\label{sec:conclusions}
In this work, we explored active learning for semiconductor defect segmentation. We proposed using contrastive pretraining for initializing the segmentation model, and proposed a rareness-aware acquisition function to prioritize samples containing minority classes for labelling. When benchmarked on a semiconductor dataset composed of XRM scans of logic and memory dies, our method achieved state-of-art-performance with only $4.9\%$ labels needed to obtain $98\%$ of the performance achievable by a fully-supervised baseline. Our work demonstrates the potential of active learning to significantly reduce data requirements for defect identification in semiconductor manufacturing.


\vfill\pagebreak

\bibliographystyle{IEEEbib}
\bibliography{references}

\end{document}